# Reasoning With Qualitative Probabilities Can Be Tractable

Moisés Goldszmidt    Judea Pearl
< moises@cs.ucla.edu >    < judea@cs.ucla.edu >
Cognitive Systems Laboratory, Computer Science Department,
University of California, Los Angeles, CA 90024

## Abstract

We recently described a formalism for reasoning with if-then rules that are expressed with different levels of firmness [18]. The formalism interprets these rules as extreme conditional probability statements, specifying orders of magnitude of disbelief, which impose constraints over possible rankings of worlds. It was shown that, once we compute a priority function $Z^+$ on the rules, the degree to which a given query is confirmed or denied can be computed in $O(\log n)$ propositional satisfiability tests, where $n$ is the number of rules in the knowledge base. In this paper, we show that computing $Z^+$ requires $O(n^2 \times \log n)$ satisfiability tests, not an exponential number as was conjectured in [18], which reduces to polynomial complexity in the case of Horn expressions. We also show how reasoning with imprecise observations can be incorporated in our formalism and how the popular notions of *belief revision* and *epistemic entrenchment* are embodied naturally and tractably.

## 1 Introduction: Infinitesimal Probabilities, Rankings and Common Sense Reasoning

The uncertainty encountered in common sense reasoning fluctuates over an extremely wide range. For example, the probability that the new book on my desk is about astrology is less than one in a million. If however, I spot a Zodiac sign on page 1, the probability becomes close to 1, say 0.999. Intelligent agents are expected to reason with such rare eventualities and to produce explanations and actions whenever these occur. Given this wide range of uncertainty fluctuations and the fact that the majority of everyday decisions involve relatively low payoffs, the full precision of probability calculus may not be necessary, and an order-of-magnitude approximation may be sufficient. Thus, instead of measuring probabilities on a scale from zero to one, we can imagine projecting probability measures onto a quantized logarithmic scale and then treating beliefs that map onto two different quanta as being of different orders of magnitude.

This method of approximation gives rise to a semi-qualitative calculus of uncertainty, one in which degrees of (dis)belief are ranked by non-negative integers (corresponding perhaps to linguistic quantifiers such as "believable," "unlikely," "very rare") still capable of accounting for retraction and restoration of beliefs by Bayesian conditionalization. The origin of this approximation can be traced back to Ernest Adams [1], who developed a logic of conditionals based on infinitesimal probabilities, and to the Ordinal Condition Functions of Spohn [29]. Potential applications in nonmonotonic reasoning were noted in [21, 23] and further developed in [20, 16, 25, 17, 18, 8].

A simple way of viewing infinitesimal probabilities is to consider an ordinary probability function $P$ defined over a set $\Omega$ of possible worlds (or states of the world) $\omega$ and to imagine that the probability $P(\omega)$ is a polynomial function of some infinitesimal parameter $\varepsilon$, arbitrarily close to, yet bigger than zero; for example $1 - c_1\varepsilon$ or $\varepsilon^2 - c_2\varepsilon^4$. Accordingly, the probabilities assigned to any subset of $\Omega$ represented by a logical formula $\varphi$, as well as all conditional probabilities $P(\psi|\varphi)$, will be rational functions of $\varepsilon$. We define the ranking function $\kappa(\psi|\varphi)$ [1] as the power of the most significant $\varepsilon$-term in the expansion of $P(\psi|\varphi)$. In other words, $\kappa(\psi|\varphi) = n$ iff $P(\psi|\varphi)$ has the same order of magnitude as $\varepsilon^n$.

The following properties of ranking functions (left-hand side below) reflect, on a logarithmic scale, the usual properties of probability functions (right-hand side), with "min" replacing addition, and addition replacing multiplication:

$$\kappa(\varphi) = \min_{\omega \models \varphi} \kappa(\omega) \quad : \quad P(\varphi) = \sum_{\omega \models \varphi} P(\omega) \quad (1)$$

$$\kappa(\varphi) = 0 \text{ or } \kappa(\neg\varphi) = 0 \quad : \quad P(\varphi) + P(\neg\varphi) = 1 \quad (2)$$

---

[1]Spohn [29] called this function a "non–probabilistic" Ordinal Condition Function.

$$\kappa(\psi \wedge \varphi) = \kappa(\psi|\varphi) + \kappa(\varphi) \quad : \quad P(\psi \wedge \varphi) = P(\psi|\varphi)P(\varphi) \tag{3}$$

Parameterizing a probability measure by $\varepsilon$ and extracting the lowest exponent of $\varepsilon$ as the measure of (dis)belief is proposed as a model of the process by which people abstract qualitative beliefs from numerical probabilities and accept them as tentative truths. For example, we can make the following correspondence between linguistic quantifiers and $\varepsilon^n$:

| | | |
|---|---|---|
| $P(\phi) = \varepsilon^0$ | $\phi$ and $\neg \phi$ are believable | $\kappa(\phi) = 0$ |
| $P(\phi) = \varepsilon^1$ | $\neg \phi$ is believed | $\kappa(\phi) = 1$ |
| $P(\phi) = \varepsilon^2$ | $\neg \phi$ is strongly believed | $\kappa(\phi) = 2$ |
| $P(\phi) = \varepsilon^3$ | $\neg \phi$ is very strongly believed | $\kappa(\phi) = 3$ |
| $\vdots$ | $\vdots$ | $\vdots$ |

These approximations yield a probabilistically sound calculus, employing integer addition, for manipulating the orders of magnitude of disbelief. The resulting formalism is governed by the following principles:

1. Each world is ranked by a non-negative integer $\kappa$ representing the degree of surprise associated with finding such a world.

2. Each well-formed formula (wff) is given the rank of the world with the lowest $\kappa$ (most normal world) that satisfies that formula.

3. Given a collection of facts $\phi$, we say that $\sigma$ follows from $\phi$ with strength $\delta$ if $\kappa(\sigma|\phi) > \delta$, or, equivalently, if the $\kappa$ rank of $\phi \wedge \neg \sigma$ is at least $\delta + 1$ degrees above that of $\phi \wedge \sigma$.

Principles 1 and 2 follow immediately from the semantics described above. Principle 3 says that $\sigma$ is plausible given $\phi$ iff $P(\sigma|\phi) \geq 1 - c\varepsilon^\delta$, where $P$ is the $\varepsilon$-parametrized probability associated with that particular ranking $\kappa$. This abstraction of probabilities matches the notion of *plain belief* in what it is deductively closed;[2] the price we pay, however, is that many small probabilities do not accumulate into a strong argument (as in the lottery paradox).

The basic $\kappa$ ranking system, as described in Spohn [29], requires the specification of a complete ranking function before reasoning can commence. In other words, the knowledge base must be sufficiently rich to define the $\kappa$ associated with every world $\omega$. Unfortunately, in practice, such specification is not readily available. For example, we might be given the information that "birds normally fly" (written $\kappa(\neg f|b) > 0$) and no information whatsoever about the flying habits of red birds or non-birds. We still would like to conclude that red birds normally fly, even though the information given is not sufficient for defining a complete ranking function. In order to draw plausible conclusions from such fragmentary pieces of information, we require additional inferential machinery that should accomplish two functions: First, it should enrich the specification of the ranking function with the needed information and, second, it should operate directly on the specification sentences in the knowledge base, rather than on the rankings of worlds (which are too numerous to list). Such machinery is provided by a formalism called system-$Z^+$ [18] which accepts knowledge in the form of quantified if-then rules (e.g., "birds fly (with strength $\delta$)") and computes the plausibility of any given query (e.g., "Tim, being a red-bird, flies (with degree $\delta'$)") by syntactic manipulation of these rules.

To accomplish these functions, system-$Z^+$ incorporates two principles in addition to those given above:

4. Each input rule "if $\varphi$ then $\psi$ (with strength $\delta$)," written $\varphi \xrightarrow{\delta} \psi$, is interpreted as a constraint on the ranking $\kappa$, forcing every world in $\varphi \wedge \neg\psi$ to rank at least $\delta + 1$ degrees above the most normal world in $\varphi$, that is, $\kappa(\psi|\varphi) > \delta$.

5. Out of all rankings satisfying the constraints above, we adopt the (unique) ranking $\kappa^+$ that assigns each world the lowest possible (most normal) rank.

Principle 4 corresponds to the notion of consistency in Def. 2.2, and Principle 5 establishes that the information in the set of if-then rules (the knowledge base) is completed by assigning the maximum likelihood possible to each world allowed by the consistency constraints imposed by these rules (see Def. 2.3 and [18]). The first contribution of this paper is to improve the inference process of system-$Z^+$ and establish its tractability. A key step in the procedures developed in [18] was the computation of a priority ranking $Z^+$ on the rules in the knowledge base, which was conjectured to be intractable. In Section 3 (after some preliminary definitions in Section 2), we present a procedure for computing $Z^+$ that requires a polynomial number of propositional satisfiability tests and hence is tractable in applications permitting restricted languages, such as Horn expressions, network theories, or acyclic databases.

The second contribution of this paper is to equip system-$Z^+$ with the capability to reason with *soft* evidence or imprecise observations (Section 4). Such a capability is important when we wish to assess the plausibility of $\sigma$ (using Principle 3 above) but the context $\phi$ is not given with absolute certainty. In other words, there is some vague testimony supporting $\phi$ but that testimony is undisclosed (or cannot be articulated using the basic propositions in our language, e.g., testimony of the senses); all that can be ascertained is a summary of that testimony saying that "$\phi$ is supported to a degree $n$." We propose two different strategies for computing a new ranking $\kappa'$ from an initial one $\kappa$, given soft evidential report supporting a wff $\phi$. The first strategy, named J-conditionalization, is based on *Jeffrey's Rule of Conditioning* [24]. It interprets the report as specifying that "all things considered," the new degree of disbelief for $\neg\phi$ should be $\kappa'(\neg\phi) = n$. The second strategy, named L-conditionalization, is based on the *virtual evidence* proposal described in [23]. It interprets the report as specifying the desired *shift* in the degree of belief in $\phi$, as warranted by that report alone and "nothing else considered." We show that L-conditionalization has roughly the same complexity as ordinary conditionalization, and then we relate our formalism to the theory of belief revision in [2]. Finally, Section 5 summarizes the main results.

---

[2]If $A$ is believed and $B$ is believed then $A \wedge B$ is believed. Note that this deviates form the threshold conception of belief: if both $P(A)$ and $P(B)$ are above a certain threshold, $P(A \wedge B)$ may still be below that same threshold.

## 2   Preliminary Definitions: The Ranking $\kappa^+$

We start with a set of rules $\Delta = \{r_i \mid r_i = \varphi_i \xrightarrow{\delta_i} \psi_i, 1 \leq i \leq n\}$, where $\varphi_i$ and $\psi_i$ are propositional formulas over a finite language of atomic propositions, "$\rightarrow$" denotes a new connective, and $\delta_i$ is a non-negative integer. A truth valuation $\omega$ of the atomic propositions in the language will be called a *world*. The satisfaction of a wff $\varphi$ by a world $\omega$ is defined as usual and denoted by $\omega \models \varphi$. $\omega$ is a *model* for $\varphi$ if $\omega$ satisfies $\varphi$. Let $\Omega$ stand for the set of possible worlds. *Ranking functions* are defined as follows:

**Definition 2.1 (Rankings)** A ranking function $\kappa$ is an assignment of non-negative integers to the elements in $\Omega$, such that $\kappa(\omega) = 0$ for at least one world $\omega \in \Omega$. We extend this definition to induce rankings on wffs:

$$\kappa(\varphi) = \begin{cases} \min_{\omega \models \varphi} \kappa(\omega) & \text{if } \varphi \text{ is satisfiable,} \\ \infty & \text{otherwise.} \end{cases} \quad (4)$$

Similarly, given two wffs $\varphi$ and $\psi$ such that $\varphi$ is satisfiable, we define the conditional ranking $\kappa(\psi|\varphi)$ as $\kappa(\psi|\varphi) = \kappa(\psi \wedge \varphi) - \kappa(\varphi)$.

**Definition 2.2 (Consistency)** A ranking $\kappa$ is said to be **admissible** relative to a given $\Delta$, iff

$$\kappa(\varphi_i \wedge \psi_i) + \delta_i < \kappa(\varphi_i \wedge \neg\psi_i) \quad (5)$$

(equivalently $\kappa(\neg\psi_i|\varphi_i) > \delta_i$) for every rule $\varphi_i \xrightarrow{\delta_i} \psi_i \in \Delta$. A set $\Delta$ is **consistent** iff there exists an admissible ranking $\kappa$ relative to $\Delta$.

Consistency can be decided in $O(|\Delta|^2)$ satisfiability tests, and it is independent of the $\delta$-values assigned to the rules in $\Delta$ [18]. Eq. 5 echoes the usual interpretation of *defaults* rules [28], according to which $\psi$ holds in all *minimal* models for $\varphi$. In our case, minimality is reflected in having the lowest rank, that is the highest possible likelihood. If we say that $\omega$ *falsifies* or *violates* a rule $\varphi \xrightarrow{\delta} \psi$ whenever $\omega \models \varphi \wedge \neg\psi$, the parameter $\delta$ can be interpreted as the minimal degree of surprise (or abnormality) associated with finding the rule $\varphi \xrightarrow{\delta} \psi$ violated, given that we know $\varphi$. In probabilistic terms, consistency guarantees that for every $\varepsilon > 0$, there exists a probability distribution $P$ such that if $\varphi_i \xrightarrow{\delta_i} \psi_i \in \Delta$, then $P(\psi_i|\varphi_i) \geq 1 - c\varepsilon^{\delta_i}$.

The distinguished ranking $\kappa^+$, defined below, assigns to each world the lowest possible rank permitted by the admissibility constraints of Eq. 5 (Def. 2.2). Such an assignment reflects the assumption that, unless we are forced to do otherwise, each world is considered as normal (likely) as possible.

**Definition 2.3 (The ranking $\kappa^+$)** Let $\Delta = \{r_i \mid r_i = \varphi_i \xrightarrow{\delta_i} \psi_i\}$ be a consistent set of rules. $\kappa^+$ is defined as follows:

$$\kappa^+(\omega) = \begin{cases} 0 & \text{if } \omega \text{ does not falsify any rule in } \Delta, \\ \max_{\omega \models \varphi_i \wedge \neg\psi_i}[Z^+(r_i)] + 1 & \text{otherwise,} \end{cases} \quad (6)$$

where $Z^+(r_i)$ is a *priority* ranking on rules, defined by

$$Z^+(r_i) = \min_{\omega \models \varphi_i \wedge \psi_i}[\kappa^+(\omega)] + \delta_i. \quad (7)$$

The recursive nature between Eqs. 6 and 7 is benign, and we present an effective procedure, Procedure Z_rank, for computing $Z^+$ in the next section. In [18], we also show that Eqs. 6 and 7 define a unique admissible ranking function $\kappa^+$ that is minimal in the following sense: Any other admissible ranking function must assign a higher ranking to at least one world and a lower ranking to none.

An alternative mechanism for enriching the original specification of a ranking (probability) function in the form of a set of conditional if–then rules is studied in [17], where Maximum Entropy principle is used to select a privileged distribution among those probability distributions complying with the constraints imposed by the rules. In the language of rankings, this distribution can be represented as a set of recursive equations similar to Eqs. 6 and 7:[3]

$$\kappa^*(\omega) = \begin{cases} 0 & \text{if } \omega \text{ does not falsify any rule in } \Delta, \\ \sum_{\omega \models \varphi_i \wedge \neg\psi_i}[Z^*(r_i)] + 1 & \text{otherwise.} \end{cases} \quad (8)$$

The computation of the $Z^*$ priorities and the query-answering procedures for the maximum entropy approach has been proven to be NP-hard even for Horn clauses (see [4]).

## 3   Plausible Conclusions: Computing the $Z^+$-rank

Given a set $\Delta$, each admissible ranking $\kappa$ induces a consequence relation $\vdash_\kappa$, where $\phi \vdash_\kappa \sigma$ iff $\kappa(\sigma \wedge \phi) < \kappa(\neg\sigma \wedge \phi)$. A straightforward way to declare $\sigma$ as a plausible conclusion of $\Delta$ given $\phi$ would be to require $\phi \vdash_\kappa \sigma$ in all $\kappa$ admissible with $\Delta$. This leads to an entailment relation called $\varepsilon$-semantics [23], 0-entailment [25], and r-entailment [20], which is recognized as being too conservative. The approach we take here is to select a distinguished admissible ranking, in our case $\kappa^+$, and declare $\sigma$ as a plausible conclusion of $\Delta$ given $\phi$, written $\phi \mathrel{\vdash\!\sim} \gamma$, iff $\kappa^+(\phi \wedge \sigma^-) < \kappa^+(\phi \wedge \neg\sigma)$.[4] According to Eq. 6 in Def. 2.3, both $\kappa^+$ and $\mathrel{\vdash\!\sim}$ can be computed effectively once the priority ranking $Z^+$ on rules is known. We next present a procedure for computing $Z^+$, which is identical to the one presented in [18] save for the crucial computation of Eq. 9 (Step 3(b)). Whereas in [18] this computation was thought to require an exponential search over worlds, we now show that it can be accomplished in $O(|\Delta| \times \log |\Delta|)$ satisfiability tests. Some of the steps in Procedure Z_rank depend upon the notion of *toleration*. A rule $\phi \xrightarrow{\delta} \sigma$ is *tolerated* by $\Delta$ if the wff $\phi \wedge \sigma \bigwedge_i \varphi_i \supset \psi_i$ is satisfiable (where $i$ ranges over all rules in $\Delta$).[5]

---

[3]The equation for $Z^*$ is identical to that of $Z^+$ (with $\kappa^+$ replaced with $\kappa^*$).

## Procedure Z_rank

**Input:** A consistent knowledge base $\Delta$. **Output:** $Z^+$-ranking on rules.

1. Let $\Delta_0$ be the set of rules tolerated by $\Delta$, and let $\mathcal{R}Z^+$ be an empty set.

2. For each rule $r_i = \varphi_i \xrightarrow{\delta_i} \psi_i \in \Delta_0$, do: set $Z(r_i) = \delta_i$ and $\mathcal{R}Z^+ = \mathcal{R}Z^+ \cup \{r_i\}$.

3. While $\mathcal{R}Z^+ \neq \Delta$, do:

   (a) Let $\Delta^+$ be the set of rules in $\Delta' = \Delta - \mathcal{R}Z^+$ tolerated by $\Delta'$

   (b) For each $r : \phi \xrightarrow{\delta} \sigma \in \Delta^+$, let $\Omega_r$ denote the set of models for $\phi \wedge \sigma$ that do not violate any rule in $\Delta'$: compute

   $$Z(r) = \min_{\omega_r \in \Omega_r} [\kappa(\omega_r)] + \delta \qquad (9)$$

   where $\kappa(\omega_r) =$

   $$\max_{r_j \in \mathcal{R}Z^+} \{Z(r_j) \mid \omega_r \models \varphi_j \wedge \neg\psi_j\} + 1 \quad (10)$$

   and $r_j : \varphi_j \xrightarrow{\delta_j} \psi_j \in \mathcal{R}Z^+$.[6]

   (c) Let $r^*$ be a rule in $\Delta^+$ having the lowest $Z$; set $\mathcal{R}Z^+ = \mathcal{R}Z^+ \cup \{r^*\}$.

## End Procedure

Theorem 3.1 establishes the correctness of Procedure Z_rank, while Lemmas 3.2 and 3.3 and Theorem 3.4 determine its (polynomial) complexity.

---

[4]If we are concerned with the strength $\delta$ with which the conclusion is endorsed, then $\phi \mathrel{\vdash\!\sim}^{\delta} \gamma$ iff $\kappa^+(\phi \wedge \sigma) + \delta < \kappa^+(\phi \wedge \neg\sigma)$.

[5]The notion of toleration is also crucial for deciding consistency: $\Delta$ is consistent iff there is a tolerated default rule in every nonempty subset $\Delta'$ of $\Delta$ (Theorem 1, [18]).

[6]Note that Eqs. 9 and 10 correspond to Eqs. 7 and 6 in Def. 2.3.

**Theorem 3.1 ([18])** *The function $Z$ computed by Z_rank complies with Def. 2.3, that is $Z = Z^+$.*

**Lemma 3.2** *Let $\Delta = \{r_i \mid r_i = \varphi_i \xrightarrow{\delta_i} \psi_i\}$ be a consistent set where the rules are sorted in nondecreasing order according to priorities $Z(r_i)$. Let $\kappa(\omega)$ be defined as in Eq. 6:*

$$\kappa(\omega) = \begin{cases} 0 & \text{if } \omega \text{ does not falsify any rule in } \Delta, \\ \max_{\omega \models \varphi_i \wedge \neg\psi_i}[Z(r_i)] + 1 & \text{otherwise.} \end{cases} \quad (11)$$

*Then, for any wff $\phi$, $\kappa(\phi)$ can be computed in $O(\log|\Delta|)$ propositional satisfiability tests.*

The idea is to perform a binary search on $\Delta$ to find the lowest $Z(r)$ such that there is a model for $\phi$ that does not violate any rule $r'$ with priority $Z(r') \geq Z(r)$. This is done by dividing $\Delta$ into two roughly equal sections: top-half ($r_{mid}$ to $r_{high}$) and bottom-half ($r_{low}$ to $r_{mid}$). A satisfiability test on the wff $\alpha = \phi \bigwedge_{j=mid}^{j=n} \varphi_j \supset \psi_j$ decides on whether the search should continue (in a recursive fashion) on the bottom-half or on the top-half.[7]

**Lemma 3.3** *The value of $Z(\phi \xrightarrow{\delta} \sigma)$ in Eq. 9 can be computed in $O(\log |\mathcal{R}Z^+|)$ satisfiability tests.*

Let $\Delta'$ in Step 3(a) be equal to $\{\varphi_i \xrightarrow{\delta_i} \psi_i\}$; the computation of Eq. 9 is equivalent to computing the $\kappa$ of the wff $\sigma \wedge \phi \bigwedge_i \varphi_i \supset \psi_i$ where $i$ ranges over all the rules in $\Delta'$, by performing the binary search on the set $\mathcal{R}Z^+$.

**Theorem 3.4** *Given a consistent $\Delta$, the computation of the ranking $Z^+$ requires $O(|\Delta|^2 \times \log |\Delta|)$ satisfiability tests.*

Computing Eq. 9 in Step 3(b) can be done in $O(\log |\mathcal{R}Z^+|)$ satisfiability tests according to Lemma 3.3,[8] and since it will be executed at most $O(|\Delta|)$ times, it requires a total of $O(|\Delta| \times \log|\Delta|)$. Loop 3 is performed at most $|\Delta| - |\Delta_0|$ times, hence the whole computation of the priorities $Z^+$ on rules requires a total of $O(|\Delta|^2 \times \log |\Delta|)$ satisfiability tests.[9]

Once $Z^+$ is known, determining the strength $\delta$ with which an arbitrary query $\sigma$ is confirmed, given the information $\phi$, requires $O(\log|\Delta|)$ satisfiability tests:

---

[7]For reasons of space, formal proofs of Lemmas 3.2 and 3.3, and Theorem 3.4 can be found in the Technical Report available by request.

[8]Note that we need $\mathcal{R}Z^+$ to be sorted, nondecreasingly, with respect to the priorities $Z$. This requires that the initial values inserted in $\mathcal{R}Z^+$ in Step 2 of Procedure Z_rank be sorted — $O(|\Delta_0|^2)$ data comparisons — and that the new $Z$-value in Step 3(c) be inserted in the right place — $O(|\mathcal{R}Z^+|)$ data comparisons. We are assuming that the cost of each of these operations is much less than that of a satisfiability test.

[9]The complexity of the remaining steps in the procedure is bounded by $O(|\Delta|)$ satisfiability tests.

First $\kappa^+(\phi \wedge \sigma)$ and $\kappa^+(\phi \wedge \neg\sigma)$ are computed, using a binary search as in Lemma 3.2. Then, these two values are compared and the difference is equated with the strength $\delta$. Clearly, if the rules in $\Delta$ are of Horn form, computing the priority ranking $Z^+$ and deciding the plausibility of queries $(\phi \mathrel{\mathop{\kern0pt\vdash}\limits^{\delta}} \sigma)$ can be done in polynomial time [10].

## 4    Belief Change, Soft Evidence, and Imprecise Observations

So far, a query $\phi \mathrel{\mathop{\kern0pt\vdash}\limits^{\delta}} \sigma$ was defined as a pair of Boolean formulae $(\phi, \sigma)$, where $\phi$ (the *context*) stands for the set of observations at hand and $\sigma$ (the *target*) stands for the conclusion whose belief we wish to confirm, deny, or assess. A query $(\phi, \sigma)$ would be answered in the affirmative if $\sigma$ was found to hold in all minimally ranked models of $\phi$, and the degree of belief in $\sigma$ would be given by $\kappa(\neg\sigma \wedge \phi) - \kappa(\sigma \wedge \phi)$.

In many cases, however, the queries we wish to answer cannot be cast in such a format, because our set of observations is not precise enough to be articulated as a crisp Boolean formula. For example, assume that we are throwing a formal party and our friends Mary and Bill are invited. However, judging form their previous behavior, we believe that "if Mary goes to the party, Bill will stay home (with strength $\delta$)," written $M \mathrel{\mathop{\kern0pt\vdash}\limits^{\delta}} \neg B$. Now assume that we have a strong hunch (with degree $K$) that Mary will go to the party (perhaps because she is extremely well dressed and is not consulting the movie section in the *Times*) and we wish to inquire whether Bill will stay home. It would be inappropriate to query the system with the pair $(M, \neg B)$, because the context $M$ has not been established beyond doubt. The difference could be critical if we have arguments against "Bill staying home," for example, that he was seen renting a tuxedo. A flexible system should allow the user to assign a degree of belief to each observational proposition in the context $\phi$ and proceed with analyzing their rational consequences. Thus, a query should consist of a tuple like $(\phi_1, K_1; \phi_2, K_2; \ldots, \phi_m, K_m : \sigma)$, where each $K_i$ measures the degree to which the contextual proposition $\phi_i$ is supported by evidence.[10]

At first glance it might seems that such facility is automatically provided by system-$Z^+$, through the use of variable–strength rules. For example, to express the fact that Mary is believed to be going to the party, we can perhaps use a *dummy* rule $Obs_1 \mathrel{\mathop{\kern0pt\vdash}\limits^{K}} M$ (stating that if Mary meets the set of observations $Obs_1$ then Mary is believed to be going to the party) and then add the proposition $Obs_1$ to the context part of the query, to indicate that $Obs_1$ has taken place.

This proposal has several shortcomings, however. First, the net impact of our new rule $Obs_1 \mathrel{\mathop{\kern0pt\vdash}\limits^{K}} M$ would be sensitive to previously collected information about Mary's intentions (say she has bought a plane ticket) that we may wish to suppress. In other words, we often wish to express the assessment that, **all things considered**, Mary's going to the party is believed to degree $K$.

Second, in many systems it is convenient to treat if–then rules as a stable part of our knowledge, unperturbed by observations made about a particular individual or in any specific set of circumstances. This permits us to compile rules into a structure that allows efficient query processing. Adding query-induced rules to the knowledge base will neutralize this facility.

Finally, rules and observations combine differently: The latter should accumulate the former do not. For example, if we have two rules $a \mathrel{\mathop{\kern0pt\vdash}\limits^{\delta_1}} c$ and $b \mathrel{\mathop{\kern0pt\vdash}\limits^{\delta_2}} c$ and we observe $a$ and $b$, system-$Z^+$ would believe $c$ to a degree $\max(\delta_1, \delta_2)$. However, if $a$ and $b$ provide two independent reasons for believing $c$, the two observations together should endow $c$ with a belief that is stronger than any one component in isolation. To incorporate such cumulative pooling of evidence, we must encode the assumption that $a$ and $b$ are conditionally independent (given $c$), which is not automatically embodied in system-$Z^+$.[11]

To avoid these complications, the method we propose treats imprecise observations by invoking specialized conditioning operators, unconstrained by a rule's semantics. We distinguish between two types of evidential reports:

1. Type-J: "All things considered," our current belief in $\phi$ should become $J$.

2. Type-L: "Nothing else considered," our current belief in $\phi$ should shift by $L$.

### 4.1    Type-J: All Things Considered

Let $\phi$ be the wff representing the event whose belief we wish to update so that $\kappa'(\neg\phi) = J$ (and, consequently, $\kappa'(\phi) = 0$).[12] In order to compute $\kappa'(\psi)$ for every wff $\psi$, we rely upon *Jeffrey's Rule of Conditioning* [24]. Jeffrey's rule is based on the assumption that while the observation changed the agent's degree of belief in $\phi$ and in certain other proposition, it did not change the **conditional degree of belief** in any propositions on the evidence $\phi$ or on the evidence $\neg\phi$ [24]. Thus, letting $P'$ denote the agent's probability distribution after the report on the value of $P'(\phi)$ is incorporated,

---

[10] We remark that evidence in this paper is regarded as setting the context of a query and not as a modifier of the knowledge in $\Delta$. Statistical methods for the later task are explored in [3].

[11] The assumptions of conditional independence among converging rules is embodied in the formalism of Maximum Entropy [17].

[12] This is an immediate consequence of the semantics for rankings and corresponds to the normalization in probability theory (see Eq. 2).

and, using $P$ to denote the agent's probability distribution prior to this report, we have[13]

$$P'(\psi|\phi) = P(\psi|\phi) \text{ and } P'(\psi|\neg\phi) = P(\psi|\neg\phi), \quad (12)$$

which leads to Jeffrey's rule,

$$P'(\psi) = P(\psi|\phi)P'(\phi) + P(\psi|\neg\phi)P'(\neg\phi). \quad (13)$$

Translated into the language of rankings (using Eqs. 1–3), Eq. 13 yields

$$\kappa'(\psi) = \min[\kappa(\psi|\phi) + \kappa'(\phi); \kappa(\psi|\neg\phi) + \kappa'(\neg\phi)], \quad (14)$$

which offers a convenient way of computing $\kappa'(\psi)$ once we specify $\kappa'(\phi) = 0$ and $\kappa'(\neg\phi) = J$. Eq. 14 assumes the an especially attractive form when computing the $\kappa'$ of a world $\omega$:

$$\kappa'(\omega) = \begin{cases} \kappa(\omega|\phi) + \kappa'(\phi) & \text{if } \omega \models \phi \\ \kappa(\omega|\neg\phi) + \kappa'(\neg\phi) & \text{if } \omega \models \neg\phi \end{cases} \quad (15)$$

Eq. 15 corresponds exactly to the $\alpha$-conditionalization proposed in Spohn [29] (Def. 6, p. 117), with $\alpha = J$. If $\kappa'(\neg\phi) = \infty$, this process is equivalent to ordinary Bayesian conditionalization, since $\kappa'(\omega) = \kappa(\omega|\phi)$ if $\omega \models \phi$ and $\kappa'(\omega) = \infty$ otherwise. Note, however, that in general this conditionalization is not commutative; if $\phi_1$ and $\phi_2$ are mutually dependent (i.e., $\kappa(\phi_2|\phi_1) \neq \kappa(\phi_2)$),[14] the order in which we establish $\kappa(\neg\phi_1) = J_1$ and $\kappa(\neg\phi_2) = J_2$ might make a difference in our final belief state, represented by the ranking $k''$.[15]

## 4.2  Type-L Reports: Nothing Else Considered

L-conditionalization is appropriate for evidential reports of the type "a new evidence was obtained which, by its own merit, would *support* $\phi$ to degree $L$." Unlike J-conditionalization, the degree $L$ now specifies *changes* in the belief of $\phi$, not the absolute value of the final belief in $\phi$. As in the case of type-J reports, we assume that in naming $\phi$ as the direct beneficiary of the evidence, the intent is to convey the assumption of conditional independence, as formulated in Eq. 13. Next, we assume that the degree of support $L$ characterizes the likelihood-ratio $\lambda(\phi)$ associated with some undisclosed observation $Obs$, as is done in the method of *virtual conditionalization* [23]:

$$\lambda(\phi) = \frac{P(Obs|\phi)}{P(Obs|\neg\phi)}, \quad (16)$$

which governs the updates via the product rule

$$\frac{P'(\phi)}{P'(\neg\phi)} = \frac{\lambda(\phi)P(\phi)}{P(\neg\phi)}. \quad (17)$$

Translated into the language of rankings, this assumption yields

$$\kappa'(\phi) - \kappa'(\neg\phi) = \kappa(\phi) - \kappa(\neg\phi) - L \quad (18)$$

and, since either $\kappa'(\phi)$ or $\kappa'(\neg\phi)$ must be zero, we obtain

$$\kappa'(\phi) = \max[0; \kappa(\phi) - \kappa(\neg\phi) - L], \quad (19)$$
$$\kappa'(\neg\phi) = \max[0; \kappa(\neg\phi) - \kappa(\phi) + L]. \quad (20)$$

We see that the effect of L-conditionalization is to shift the degree of disbelief difference between $\phi$ and $\neg\phi$ by the specified amount $L$. Once $\kappa'(\phi)$ is known, we can use Jeffrey's rule (Eq. 14) to compute the $\kappa'(\psi)$ for an arbitrary wff $\psi$, we have that $\kappa'(\psi) =$

$$\begin{cases} \min[\kappa(\psi|\phi) + \kappa(\phi) - \kappa(\neg\phi) - L; \kappa(\psi|\neg\phi)] \\ \min[\kappa(\psi|\phi); \kappa(\psi|\neg\phi) + \kappa(\neg\phi) + L - \kappa(\phi)] \\ \min[\kappa(\psi|\phi); \kappa(\psi|\neg\phi)] \end{cases} \quad (21)$$

depending on whether $\kappa(\neg\phi) + \kappa(\phi)$ is less than, greater than, or equal to $L$. This expression takes the following form for $\kappa'(\omega)$:

$$\kappa'(\omega) = \begin{cases} \kappa(\omega|\phi) + \max[0; \kappa(\phi) - \kappa(\neg\phi) - L] \\ \kappa(\omega|\neg\phi) + \max[0; \kappa(\neg\phi) - \kappa(\phi) + L] \end{cases} \quad (22)$$

depending on whether $\omega \models \phi$ or $\omega \models \neg\phi$. As in J-conditionalization, if $L = \infty$ then $\kappa'(\omega) = \kappa(\omega|\phi)$. For the general case, we can see that the effect of L-conditionalization is to shift *downward* the $\kappa$ of all worlds that are models of the supported proposition $\phi$ relative to the $\kappa$ of all worlds that are not models for $\phi$. However, unlike J-conditionalization, the net relative shift is constant and is equal to $L$, independent of the initial value of $\kappa(\phi)$. It is easy to verify that L-conditionalization is commutative (as is its probabilistic counterpart, see Eq. 17), and hence it permits a recursive implementation in the case of multiple evidence.

We can illustrate these updating schemes through the party example consisting of the single rule $r_m : M \xrightarrow{4} \neg B$ ("if Mary goes to the party, then Bill will not go"). A trivial application of Procedure Z_rank yields $Z^+(r_m) = 4$, and using Eqs. 4 and 6 we find $\kappa(x) = 0$ for every proposition $x$, except $x = B \wedge M$, for which we have $\kappa^+(M \wedge B) = 5$. This means that we have no reason to believe that either Mary or Bill will go to the party, but we are pretty sure that both of them will not show up. Now suppose we see that Mary is very well dressed, and this observation makes our belief in M increase to 3, that is, $\kappa^{+'}(\neg M) = 3$. As a consequence, our belief in Bill staying home also increases to 3 since, using either J-conditionalization or L-conditionalization, $\kappa^{+'}(B) = 3$. Next, suppose that someone tells us that he has a strong hunch that Bill plans to show up for the party, but he fails to tell us why. There are two ways in which this report can influence our beliefs. The natural way would be to assume that our informer has not seen Mary's dress, and might not even be aware of Bill and Mary's relationship — hence we assess the impact of his report in isolation and say that whatever the value of our current belief in Bill going, it should increase by 3 increments, or $L = 3$. Following Eq. 21, $\kappa^{+''}(B)$ and $\kappa^{+''}(\neg M)$ will both be equal to 0, and we are back to the initial uncertainty about Bill or Mary going to the party, except that our disbelief in both Mary and Bill being at the party has decreased to $\kappa^{+''}(M \wedge B) = 2$. The second way would be to assume that our informer is omniscient and already has taken into consideration all we know about Bill and Mary. He means for us to revise our rankings so that the final belief in "Bill going" will be fixed at $\kappa^{+''}(\neg B) = 3$. With this interpretation, we J-condition $\kappa^{+'}$ on the proposition $\phi = \neg B$ and obtain $\kappa^{+''}(M) = 3$, concluding that it is Mary who will not show up to the party after all.

---

[13] Eq. 12 is known as the *J-condition* [24].

[14] This condition mirrors probabilistic dependence, i.e., $P(\phi_2|\phi_1) \neq P(\phi_2)$.

[15] Spohn ([29], p. 118) has acknowledged the desirability of commutativity in evidence pooling but has not stressed that $\alpha$-conditionalization commutes only in a very narrow set of circumstances (partially specified by his Theorem 4). These circumstances require that successive pieces of evidence support only propositions that are relatively independent — the truth of one proposition should not imply a belief in another. Shenoy [27] has corrected this deficiency by devising a commutative combination rule which behaves similar to L-conditioning.

## 4.3   Complexity Analysis

From Eq. 14 we see that $\kappa'(\psi)$ can be computed from $\kappa(\psi|\phi)$ and $\kappa(\psi|\neg\phi)$, which, assuming we have $Z^+$, requires a logarithmic number of propositional satisfiability tests (see Section 3). L-conditionalization can follow a similar route, as depicted in Eq. 21.

Special precautions must be taken when simultaneous, multiple pieces of evidence become available. First, J-conditionalization is not commutative, hence we cannot simply compute $\kappa'$ by J-conditioning on $\phi_1$ and then J-conditioning $\kappa'$ on $\phi_2$ to get $\kappa''$. We must J-condition simultaneously on $\phi_1$ and $\phi_2$ with their respective J-levels, say $J_1$ and $J_2$. Worse yet, an auxiliary effort must be expended to compute the J-level of each combination of $\phi$'s, in our case $\phi_1 \wedge \phi_2$, $\phi_1 \wedge \neg\phi_2$, etc. This is no doubt a hopeless computation when the number of observations is large.

L-conditionalization, by virtue of its commutativity, enjoys the benefits of recursive computations. Let $e_1$ and $e_2$ be two (undisclosed) pieces of evidence supporting $\phi_1$ (with strength $L_1$) and $\phi_2$ (with strength $L_2$), respectively. We first L-condition $\kappa$ on $\phi_1$ and calculate $\kappa'(\phi_1)$ and $\kappa'(\phi_2)$ using Eq. 20 and Eq. 21, respectively. Applying Eq. 21 this time to $\kappa'(\psi \wedge \phi_2)$, we calculate $\kappa'(\psi|\phi_2)$. Second, we L-condition $\kappa'$ on $\phi_2$, compute $\kappa''(\phi_2)$ using Eq. 20, and, finally, using $\kappa'(\psi|\phi_2)$ and $\kappa''(\phi_2)$ in Eq. 21 obtain $\kappa''(\psi)$.[16] Note that, although each of these calculations requires only $O(\log |\Delta|)$ satisfiability tests, this computation is effective only when we have a well designated target hypothesis $\psi$ to estimate. The computation must be repeated each time we change the target hypothesis, even when the context remains unaltered. This is so because we no longer have a facility for encoding a complete description of $\kappa'$, as we had for $\kappa$ (using the $Z^+$-function). Thus, the encoding for $\kappa'$ may not be as *economical* as that for $\kappa$ (the number of worlds is astronomical), unless we manage to find dummy rules that emulate the constraints imposed on $\phi_1$ by the (undisclosed) observation. Such dummy rules must enforce the conditional independence constraints embedded in Eq. 13, without violating the admissibility constraints (Eq. 5) in $\Delta$ (see [19]).

[16]The generalization to more than two pieces of evidence is straightforward.

## 4.4   Relation to the AGM Theory of Revision

Alchourrón, Gärdenfors and Makinson (AGM) have advanced a set of postulates that have become a standard against which proposals for belief revision are tested [2]. The AGM approach models epistemic states as deductively closed sets of (believed) sentences and characterizes how a rational agent should change its epistemic states when new beliefs are added, substracted, or changed. The central result of this theory is that these postulates are equivalent to the existence of a complete preordering of all propositions according to their degree of *epistemic entrenchment* such that belief revisions always retain more entrenched propositions in preference to less entrenched ones. However, the AGM postulates do not provide a calculus with which one can realize the revision process or even specify the content of an epistemic state [5, 11, 22].

Spohn [29] has shown how belief revision conforming to the AGM postulates can be embodied in the context of ranking functions. Once we specify a single ranking function $\kappa$ on possible worlds, we can associate the set of beliefs with those propositions $\psi$ for which $\kappa(\neg\psi) > 0$. To incorporate a new belief $\beta$, one can raise the $\kappa$ of all models of $\neg\beta$ relative to those of $\beta$, until $k(\neg\beta)$ becomes (at least) 1, at which point the newly shifted ranking defines a new set of beliefs. This process of belief revision, corresponding to $\alpha$-conditioning (with $\alpha = 1$) was shown to obey the AGM postulates, from which it follows that revision in the probabilistic system described in this paper also obeys those postulates under the same interpretation of beliefs.

Still, neither the AGM theory nor Spohn's embodiment of the theory are directly applicable to AI systems, the former because it does not provide a finite specification for belief sets and their entrenchment ordering, and the latter because it requires an explicit encoding of the ranking function on all possible worlds.

To better model AI practice, Nebel [22] adapted the AGM theory so that finite sets of *base* propositions mediate revisions. This is exemplified in the nonmonotonic systems of Brewka [7] and Poole [26], where the base consists of propositional implications, or *expectations* [15]. The basic idea in these syntax-based systems is to define a (total) priority order on the set of base propositions, and select revisions to be maximally consistent relative to that order. Nebel has shown that such a strategy can satisfy almost all the AGM ax-ioms. Boutilier [5] has further shown that, indeed, the priority function $Z^+$ corresponds naturally to the epistemic entrenchment ordering of the AGM theory. [17]

Unfortunately, even Nebel's theory does not completely succeed at formalizing the practice of belief revision, as it does not specify how the priority order on the base propositions is to be determined. Although one can imagine, in principle, that the knowledge author specify this order in advance, such specification would be impractical, since the order might change whenever new rules are added to the knowledge base.

Thus we see that there are several computational and epistemological advantages to our system over those proposed by AGM and Spohn, stemming from the fact that our revision process revolves around a finite set of conditional rules, not around the beliefs, the rankings or the expectations that emanate from those rules. First, since the number of propositions in one's belief set is astronomical, and so is the number of worlds, it is a computational necessity to base belief revision on rules, whose number is usually manageable. Second, our system extracts both beliefs and rankings of beliefs automatically from the content of $\Delta$; no outside specification of belief orderings is required.

Third, in order to facilitate recovery from obsolete observations Spohn's framework must assume that all observations are defeasible (or imprecise), which corresponds to $\alpha$-conditioning with $\alpha < \infty$. In our system, we can accommodate both imprecise and precise observations (corresponding to $\alpha = \infty$) using ordinary conditioning. Given a set of precise observations $\phi$, the set of beliefs is defined as those propositions $\sigma$ for which $\kappa(\neg\sigma|\phi) > 0$. Retraction of obsolete observations can be done by simply removing those observations from $\phi$. This flexibility is facilitated by maintaining a fixed set of conditional rules, with the help of which one can always restore beliefs (and rankings) so that they reflect the observations at hand, independently of those seen in the past.

Finally, and perhaps most significantly, our system is capable of responding not merely to empirical observations but also to linguistically transmitted information such as conditional sentences (i.e., if–then rules). For example, suppose someone tells us that Mary too tries to avoid Bill in parties; we simply add this information to our knowledge base in the form of a new rule, $B \longrightarrow \neg M$, recompute $Z^+$, and are prepared to respond to new observations or hearsay. In Spohn's system, where revisions are limited to $\alpha$-conditioning, one cannot properly revise beliefs in response to conditional statements.[18]

Having the capability of adopting new conditionals (as rules) also provides a simple semantics for interpreting more complex sentences involving conditionals (e.g., "If you wear a helmet whenever you ride a motorcycle, then you wont get hurt badly if you fall"[19]). Both nested and negated conditionals cease to be a mystery once we permit explicit references to default rules. The sentence "If $(a \longrightarrow b)$ then $(c \longrightarrow d)$" is interpreted as:

"If I add the default $a \longrightarrow b$ to $\Delta$, then $(c, d)$ will be in the consequence relation $\vdash_{\widetilde{\kappa}}$ of the resulting knowledge base $\Delta' = \Delta \cup \{a \longrightarrow b\}$."

which is clearly a proposition that can be tested in the language of default-based ranking systems.

Note the essential distinction between having a conditional rule $a \longrightarrow b$ explicitly in $\Delta$ and certifying that $a \vdash_{\widetilde{\kappa}} b$ holds in the consequence relation of $\Delta$. The former is a permanent constraint that interacts with other rules to shape the priority order $Z^+$, while the latter indicates a contingent expectation (certified by the Ramsey test) that passively reflects that ordering. This distinction gets lost in systems such as Spohn's or Gärdenfors' [15] that do not acknowledge the centrality of conditionals as the basis for generating and ranking beliefs.[20]

## 5    Conclusions

This paper proposes a belief-revision system that reasons tractably and plausibly with linguistic quantification of both observational reports (e.g., "looks like") and domain rules (e.g., "typically"). We have shown that the system is semi-tractable, namely, tractable for every sublanguage in which propositional satisfiability is polynomial (Horn expressions, network theories, acyclic expressions, etc.). To the best of our knowledge, this is the first system that reasons with approximate probabilities which offers such broad guarantees of tractability.[21] We expect these results to carry over to the theory of possibility as formulated by Dubois and Prade [12], which has similar features to Spohn's system except that beliefs are measured on the real interval $[0, 1]$. In addition we have shown that, without loss of tractability, the system can also accommodate expressions of imprecise observations, thus pro-viding a good model for weighing the impact of evidence and counter-evidence on our beliefs. Finally, we have shown that the enterprise of belief revision, as formulated in the work presented in [2], can find a tractable and natural embodiment in system-$Z^+$, unhindered by difficulties that plagued earlier systems.

---

[18]Gardenfors [14] attempts to devise postulates for conditional sentences, but finds them incompatible with the Ramsey test (page 156-160). See also Boutilier [5] for an analysis of Ramsey test and the AGM axioms.

[19]Example due to Calabrese (personal communication).

[20]Belief revision systems proposed in the database literature [13, 9] suffer from the same shortcoming. In that context defaults (and conditionals) represent integrity constraints with exceptions.

[21]Whereas most tractability results exploit the topological structure of the knowledge base (hypertrees, or partial hypertrees), ours are topology-independent.

## Acknowledgements

This work was supported in part by National Science Foundation grant #IRI-88-21444 and State of California MICRO 90-127. The first author was supported by an IBM graduate fellowship 1990-1992.

## References

[1] E. W. Adams. *The Logic of Conditionals*. D.Reidel, Dordrecht, Netherlands, 1975.

[2] C. Alchourrón, P. Gärdenfors, and D. Makinson. On the logic of theory change: Partial meet contraction and revision functions. *Journal of Symbolic Logic*, 50:510–530, 1985.

[3] F. Bacchus. *Representing and Reasoning with Probabilistic Knowledge, A Logical Approach to Probabilities*. The MIT Press, Cambridge., 1991.

[4] R. Ben-Eliyahu. NP-complete problems in optimal horn clauses satisfiability. Technical Report R-158, Cognitive systems lab, UCLA, 1990.

[5] C. Boutilier. Conditional logics for default reasoning and belief revision. Ph.D. dissertation, University of Toronto, 1992.

[6] C. Boutilier. What is a default priority? In *Proceedings of the Canadian Conference on Artificial Intelligence*, Vancouver, 1992.

[7] G. Brewka. Preferred subtheories: An extended logical framework for default reasoning. In *Proceedings of the International Joint Conference in Artificial Intelligence (ICAI-89)*, Detroit, 1989.

[8] W. Buntine. Some properties of plausible reasoning. In *Proceedings of the $7^{th}$ Conference on Uncertainty in AI*, pages 44–51, Los Angeles., 1991.

[9] M. Dalal. Investigations into a theory of knowledge base revision: Preliminary report. In *Proceedings of the Seventh National Conference on Artificial Intelligence*, pages 475–479, 1988.

[10] W. Dowling and J. Gallier. Linear-time algorithms for testing the satisfiability of propositional Horn formulae. *Journal of Logic Programming*, 3:267–284, 1984.

[11] J. Doyle. Rational belief revision (preliminary report). In *Proceedings of Principles of Knowledge Representation and Reasoning*, pages 163–174, Cambridge. Massachusetts, 1991.

[12] D. Dubois and H. Prade. *Possibility Theory: An Approach to Computerized Processing of Uncertainty*. Plenum Press, New York, 1988.

[13] R. Fagin, J. D. Ullman, and M. Vardi. On the semantics of updates in databases. In *Proceedings of the $2^{nd}$ ACM SIGACT-SIGMOD-SIGART Symposium on Principles of Database Systems*, pages 352–365, 1983.

[14] P. Gärdenfors. *Knowledge in Flux: Modeling the Dynamics of Epistemic States*. MIT Press, Cambridge, 1988.

[15] P. Gärdenfors. Nonmonotonic inferences based on expectations: A preliminary report. In *Principles of Knowledge Representation and Reasoning: Proceedings of the Second International Conference*, pages 585–590, Boston, 1991.

[16] H. A. Geffner. Default reasoning: Causal and conditional theories. Ph.D. dissertation, University of California Los Angeles, Computer Science Department, Los Angeles, 1989. Forthcoming, MIT Press.

[17] M. Goldszmidt, P. Morris, and J. Pearl. A maximum entropy approach to nonmonotonic reasoning. In *Proceedings of American Association for Artificial Intelligence Conference*, pages 646–652, Boston, 1990.

[18] M. Goldszmidt and J. Pearl. System $Z^+$: A formalism for reasoning with variable strength defaults. In *Proceedings of American Association for Artificial Intelligence Conference*, Anaheim, CA, 1991.

[19] M. Goldszmidt and J. Pearl. Stratified rankings for causal relations. In *Proceedings of the Fourth International Workshop on Nonmonotonic Reasoning* (forthcoming), Vermont, 1992.

[20] D. Lehmann. What does a conditional knowledge base entail? In *Proceedings of Principles of Knowledge Representation and Reasoning*, pages 212–222, Toronto, 1989.

[21] J. McCarthy. Applications of circumscription to formalizing commonsense knowledge. *Artificial Intelligence*, 28:89–116, 1986.

[22] B. Nebel. Belief revision and default reasoning: Syntax-based approaches. In *Proceedings of Principles of Knowledge Representation and Reasoning*, pages 417–428, Cambridge, Massachusetts, 1991.

[23] J. Pearl. *Probabilistic Reasoning in Intelligent Systems: Networks of Plausible Inference*. Morgan Kaufmann, San Mateo, CA, 1988.

[24] J. Pearl. Jeffrey's rule, passage of experience and neo-Bayesianism. In R. Parikh, editor, *Defeasible Reasoning and Knowledge Representation*, pages 121–135. Kluwer Publishers, San Mateo, 1990.

[25] J. Pearl. System Z: A natural ordering of defaults with tractable applications to default reasoning. In R. Parikh, editor, *Proceedings of Theoretical Aspects of Reasoning about Knowledge*, pages 121–135. Morgan Kaufmann, San Mateo, CA, 1990.

[26] D. Poole. A logical framework for default reasoning. *Artificial Intelligence*, 36:27–47, 1988.

[27] P. P. Shenoy. On Spohn's rule for revision of beliefs. *International Journal of Approximate Reasoning*, 5(2):149–181, 1991.

[28] Y. Shoham. Nonmonotonic logics: Meaning and utility. In *Proceedings of International Conference on AI (IJCAI.87)*, pages 388–393, Milan, Italy, 1987.

[29] W. Spohn. Ordinal conditional functions: A dynamic theory of epistemic states. In W. L. Harper and B. Skyrms, editors, *Causation in Decision, Belief Change, and Statistics*, pages 105–134. Reidel, Dordrecht, Netherlands, 1987.